%% file: main.tex
\documentclass[letterpaper, 10 pt, conference]{ieeeconf}
\IEEEoverridecommandlockouts
\usepackage{graphicx}
\usepackage{epsfig}
\usepackage{mathptmx}
\usepackage{times}
\usepackage{amsmath}
\usepackage{amssymb}
\usepackage{diagbox}
\usepackage{url}
\usepackage[usenames]{xcolor}

\usepackage{xifthen}
\usepackage{xparse}
\usepackage{subdepth}
\usepackage{leftidx}
\usepackage{bm}

\newcommand{\bbm}{\begin{bmatrix}}
\newcommand{\ebm}{\end{bmatrix}}

\DeclareMathAlphabet{\mbf}{OT1}{ptm}{b}{n}
\newcommand{\mbs}[1]{{\bm{#1}}}

\newcommand{\mbsbar}[1]{{\overline{\boldsymbol{#1}}}}
\newcommand{\mbshat}[1]{{\hat{\boldsymbol{#1}}}}
\newcommand{\mbstilde}[1]{{\tilde{\boldsymbol{#1}}}}

\newcommand{\mbsdot}[1]{{\dot {\boldsymbol{#1}}}}

\newcommand{\mbfbar}[1]{{\overline{\mbf{#1}}}}
\newcommand{\mbfhat}[1]{{\hat{\mbf{#1}}}}
\newcommand{\mbftilde}[1]{{\tilde{\mbf{#1}}}}

\newcommand{\mbfdot}[1]{{\dot{\mbf{#1}}}}

\newcommand{\cframe}[1]{{\smash{\protect\underrightarrow{\mathcal{F}}_{#1}}}}

\DeclareMathAlphabet{\mathbfit}{OML}{cmm}{b}{it}
\newcommand{\homo}[1]{{\mathbfit{#1}}}

\newcommand{\mbfh}[1]{{\homo{#1}}}

\newcommand{\vel}[3]{\leftidx{_{#1}}{\mbf v}{\IfValueTF{#2}{_{#2#3\hspace{2pt}}}{}}} 
\newcommand{\veltilde}[3]{\leftidx{_{#1}}{\mbftilde v}{\IfValueTF{#2}{_{#2#3\hspace{2pt}}}{}}} 
\newcommand{\velbar}[3]{\leftidx{_{#1}}{\mbfbar v}{\IfValueTF{#2}{_{#2#3\hspace{2pt}}}{}}} 
\newcommand{\velhat}[3]{\leftidx{_{#1}}{\mbfhat v}{\IfValueTF{#2}{_{#2#3\hspace{2pt}}}{}}} 
\newcommand{\veldot}[3]{\leftidx{_{#1}}{\mbfdot v}{\IfValueTF{#2}{_{#2#3\hspace{2pt}}}{}}} 

\newcommand{\acc}[3]{\leftidx{_{#1}}{\mbf a}{\IfValueTF{#2}{_{#2#3\hspace{2pt}}}{}}} 
\newcommand{\acctilde}[3]{\leftidx{_{#1}}{\mbftilde a}{\IfValueTF{#2}{_{#2#3\hspace{2pt}}}{}}} 
\newcommand{\accbar}[3]{\leftidx{_{#1}}{\mbfbar a}{\IfValueTF{#2}{_{#2#3\hspace{2pt}}}{}}} 

\newcommand{\rotvel}[3]{\leftidx{_{#1}}{\mbs \omega}{\IfValueTF{#2}{_{#2#3\hspace{2pt}}}{}}} 
\newcommand{\rotveltilde}[3]{\leftidx{_{#1}}{\mbstilde \omega}{\IfValueTF{#2}{_{#2#3\hspace{2pt}}}{}}} 
\newcommand{\rotvelbar}[3]{\leftidx{_{#1}}{\mbsbar \omega}{\IfValueTF{#2}{_{#2#3\hspace{2pt}}}{}}} 
\newcommand{\rotvelhat}[3]{\leftidx{_{#1}}{\mbshat \omega}{\IfValueTF{#2}{_{#2#3\hspace{2pt}}}{}}} 
\newcommand{\rotveldot}[3]{\leftidx{_{#1}}{\mbsdot \omega}{\IfValueTF{#2}{_{#2#3\hspace{2pt}}}{}}} 

\newcommand{\T}[2]{\leftidx{}{\mbfh T}{_{#1#2\hspace{2pt}}}} 

\title{\LARGE \bf Finding Things in the Unknown: \\
Semantic Object-Centric Exploration with an MAV}

\author{
Sotiris Papatheodorou$^{1,2,3}$,
Nils Funk$^{1}$,
Dimos Tzoumanikas$^{1}$,
\\
Christopher Choi$^{1}$,
Binbin Xu$^{1,4}$
and Stefan Leutenegger$^{1,2,3}$%
\thanks{This work was funded in part by the Imperial College President's PhD Scholarship, SLAMcore Ltd., EPSRC grant ORCA Stream B - Towards Resident Robots, and EPSRC grant Aerial ABM EP/N018494/1.}%
\thanks{$^{1}$Smart Robotics Lab, Department of Computing, Imperial College London. E-mail addresses:
\texttt{\{s.papatheodorou18, nils.funk13, dimosthenis.tzoumanikas14, christopher.choi, b.xu17, s.leutenegger\}@ic.ac.uk}}
\thanks{$^{2}$Smart Robotics Lab, Department of Informatics, Technical University of Munich. E-mail address:
            \texttt{stefan.leutenegger@tum.de}}
\thanks{$^3$ Munich Institute of Robotics and Machine Intelligence (MIRMI)}
\thanks{$^4$ University of Toronto Robotics Institute, University of Toronto. E-mail address: \texttt{binbin.xu@utoronto.ca}}%
}

\begin{document}
\maketitle
\thispagestyle{empty}
\pagestyle{empty}

\begin{abstract}
	Exploration of unknown space with an autonomous mobile robot is a well-studied problem.
	In this work we broaden the scope of exploration, moving beyond the pure geometric goal of uncovering as much free space as possible.
	We believe that for many practical applications, exploration should be contextualised with semantic and object-level understanding of the environment for task-specific exploration.
	Here, we study the task of both \emph{finding} specific objects in unknown space as well as \emph{reconstructing} them to a target level of detail.
	We therefore extend our environment reconstruction to not only consist of a background map, but also object-level and semantically fused submaps.
	Importantly, we adapt our previous objective function of uncovering as much free space as possible in as little time as possible with two additional elements: first, we require a maximum observation distance of background surfaces to ensure target objects are not missed by image-based detectors because they are too small to be detected.
	Second, we require an even smaller maximum distance to the found objects in order to reconstruct them with the desired accuracy.
	We further created a Micro Aerial Vehicle (MAV) semantic exploration simulator based on Habitat in order to quantitatively demonstrate how our framework can be used to efficiently find specific objects as part of exploration.
	Finally, we showcase this capability can be deployed in real-world scenes involving our drone equipped with an Intel RealSense D455 RGB-D camera.
\end{abstract}


\section{Introduction}

Exploring an unknown environment using an autonomous robot has been well-researched.
The objective is typically formulated as uncovering as much free space as possible in a given timeframe, or uncovering all the somehow limited free space as fast as possible \cite{Yamauchi_1997CIRA,Bircher_AR2018,2020_ICRA_DPFTL,2022_RAL_Duberg}.
Hereby, a tight interaction between free space mapping, planning, and control is crucial.
MAVs are ideal platforms for autonomous exploration due to their ability to quickly move in any direction, allowing them to explore complex 3D environments.

However, exploration isn't often a goal in and of itself, it is a necessary step in solving a larger task.
We believe that many robotics tasks that require exploration also require object-level and semantic understanding of the environment.
Although there has been work aimed at exploring an unknown environments while detecting objects of interest, their goal is often finding a single object instance \cite{2020_NeurIPS_Chaplot,2020_ICRA_Zeng} or isn't focused on object reconstruction \cite{Dang_2018IEEEAC}.

In this paper we propose a task-specific exploration method aimed at finding all objects of interest in an unknown environment and reconstructing them with high accuracy.
To the best of our knowledge, this is the first work using object-level maps for active object reconstruction and exploration.
In summary, we propose the following contributions:

\begin{itemize}
	\item A mapping pipeline suitable for exploration, path planning and high quality object reconstructions.
	\item An exploration utility function tailored to discovering objects and creating high quality object reconstructions.
	\item A scheme that accounts for incomplete depth maps preventing the exploration algorithm from getting stuck in regions of the scene where depth measurements consistently cannot be obtained.
	\item An open source MAV exploration simulator based on Habitat \cite{2019_ICCV_Savva} and the Robot Operating System (ROS) \footnote[4]{https://github.com/smartroboticslab/semantic-exploration-icra-2023}.
\end{itemize}

\begin{figure}[t]
	\centering
	\includegraphics[width=0.493\columnwidth]{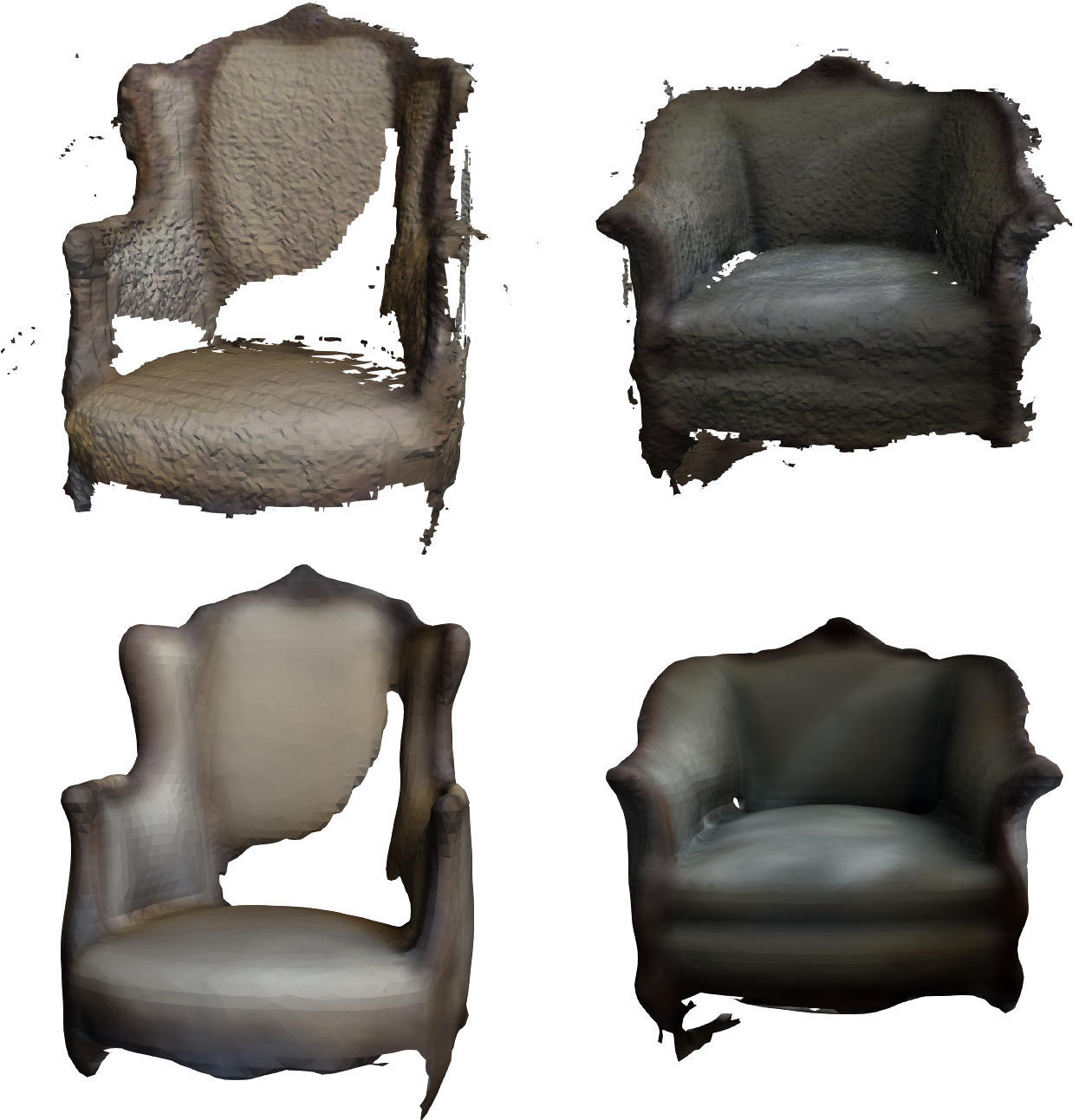}
	\includegraphics[width=0.493\columnwidth]{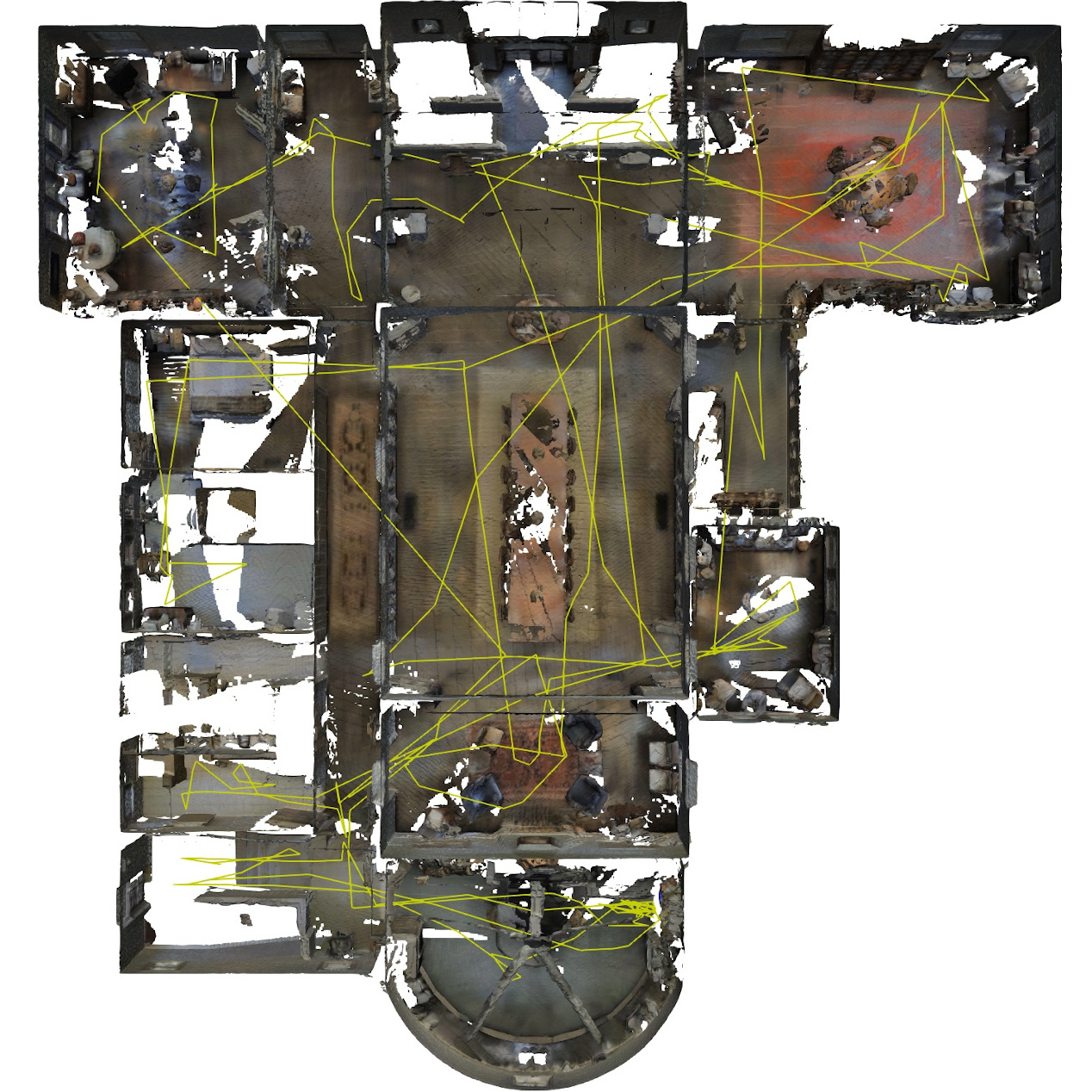}
	\caption{Objects reconstructed using our method [Top left] and their corresponding ground-truth meshes provided by the Matterport3D dataset \cite{2017_3DV_Chang} [Bottom left]. Top-down view of 3D reconstruction after exploration, and MAV path in yellow [Right].}
	\label{fig_cover}
\end{figure}
\section{Related Work}

The concept of using frontiers for exploration was introduced in \cite{Yamauchi_1997CIRA}.
Frontiers are the boundaries between known free and unknown space and indicate regions that will expand the map of known space when observed.
In \cite{Yamauchi_1997CIRA} the closest frontier was selected as the robot's next goal but various methods of selecting the best frontier to visit next have been proposed \cite{Cieslewski_2017IROS,Gao2018AnIF,Faria2019} since.

A more recent family of exploration methods, pioneered by \cite{Bircher_AR2018}, are those based on next-best-view selection.
In \cite{Bircher_AR2018} candidate next views are sampled in free space and a utility computed for each one based on the unknown volume observed from the candidate and the length of the path to the candidate.
The candidate view with the highest utility is then selected as the next robot goal.
Informed sampling, e.g.\ close to frontiers \cite{2020_ICRA_DPFTL,2021_RAL_Kompis,2022_RAL_Zhong,2022_RAL_Duberg} can greatly increase the sample-efficiency of these methods.
The choice of candidate view utility function is another important design consideration for an exploration algorithm \cite{2021_RAL_Akbari,2021_RAL_Schmid,2022_RAL_Batinovic,2022_RAL_Li}.

The SLAM community has recently moved its attention from pure geometric representations to semantically annotated reconstructions \cite{McCormac:etal:ICRA2017}.
One approach is to build object-centric maps for each detected object in the scene.
SLAM++ \cite{Salas-Moreno:etal:CVPR2013} is an early object-level mapping system that matches observed objects to pre-scanned shapes.
Later, works such as Fusion++ \cite{McCormac_3DV2018} and Kimera \cite{Rosinol:etal:ICRA2020} were proposed to incrementally build object-level maps inside a dense SLAM system.
Object-level representations in a SLAM system have shown to provide more reliable loop closure detection \cite{McCormac_3DV2018} and also robustness to dynamic objects \cite{Xu_2019ICRA}.

Powered by these advancements, several recent works propose the inclusion of semantic information for autonomous exploration in unknown environments.
In
\cite{Dang_2018IEEEAC,2020_ICRA_Zeng}, the authors propose a method to create a semantically-annotated mapping system and include the re-observation of detected objects in the utility for the proposed sampling-based path planning algorithm. 
\cite{2020_RS_Ashour} further takes inter-object spatial relationships into consideration by maintaining semantic linking maps for the next-best-view selection.
In terms of information-based exploration, \cite{Asgharivaskasi:Atanasov:ICRA2021} develops a Bayesian multi-class semantic mapping system, where a closed-form lower bound for Shannon mutual information is computed to evaluate an optimal trajectory.
Instead of using local goal detection and heuristic utility functions, learning-based approaches have also been proposed to train a navigation policy network in the semantic map via reinforcement learning \cite{2020_NeurIPS_Chaplot}.
The goal of these semantically-guided exploration methods is either finding one object of a specific class or finding all objects in the environment as a result of classic exploration.
They do not focus on reconstructing the found objects in detail nor employ an exploration algorithm to facilitate the detection of all objects in the environment.
It is these gaps that our proposed method aims to fill.
\section{Problem Formulation}

The goal of this work is to use a sensor-equipped MAV to explore and map an unknown space and, at the same time, find and reconstruct in detail an unknown number of objects.
Thus, the MAV must be able to map its environment, plan safe exploration paths, detect objects and create separate object reconstructions.

\subsection{Environment Model}

We model the static environment as a bounded volume $V \subset \mathbb{R}^3$ whose points $\mbf{v} \in V$ have an associated occupancy probability $P_o(\mbf{v})$.
If no prior information about the environment is available, the occupancy of all points $\mbf{v} \in V$ is initially \textit{unknown}, defined as $P_o(\mbf{v})=0.5$.
Due to the geometry of the environment and the MAV, as well as the sensor's mounting pose, there can be points $V_\mathrm{unob} \subset V$ that cannot be observed by the sensor.
The goal of exploration is to create a map $M$ of the observable part of the environment $V_\mathrm{obs} = V \setminus V_\mathrm{unob}$ by updating the occupancy probability of all $\mbf{v} \in V_\mathrm{obs}$ to either \textit{free} or \textit{occupied}.
Frontiers \cite{Yamauchi_1997CIRA}, the boundaries between \textit{free} and \textit{unknown} space, identify regions that will extend the map when observed.
The map $M$ can be used to plan collision-free paths for the MAV to follow.

It is assumed that the environment contains $N_o \in \mathbb{N}$ static objects of interest, each one assigned a semantic class from the set of semantic classes $\mathcal{C}$.
Each semantic class has an associated target map resolution $r_c, \, c \in \mathcal{C}$ selected by the user.
The goal of object-centric exploration is to find all objects and create a high-quality map $M_i^c, \, i \in \left\{1 \dots N_o\right\}, \, c \in \mathcal{C}$ with a resolution at least as fine as $r_c$ for each one.

\subsection{MAV Model}

For the purposes of exploration and path planning, the MAV's state $\mbf{x}$ consists of its position vector $\mbf{r} = [x, y, z]^T \in V$ in the world frame $\cframe{W}$ and a yaw angle scalar, $\psi \in [-\pi, \pi)$ with respect to the origin of the world coordinate,  thus $\mbf{x} = [x, y, z, \psi]^T \in V \times [-\pi, \pi)$.
We assume that the MAV has a maximum linear velocity $\mbf{\upsilon}_{\max} \in \mathbb{R}^+$ and a maximum yaw rate $\mbf{\omega}_{\max} \in \mathbb{R}^+$ and is enclosed in a sphere of radius $R$ centred at $\mbf{r}$.
While our planning does not consider the MAV's dynamics or roll and pitch angles, the full 6 degree of freedom pose is needed for mapping.
It is expressed as a rigid body transformation matrix $\T{W}{C}$ from the MAV camera frame $\cframe{C}$ to the world frame $\cframe{W}$ and can be estimated using an onboard SLAM system or an external motion capture system.

The MAV is equipped with an RGB-D sensor of resolution $W \times H$, focal lengths $f_x$ and $f_y$, and
depth range $\left[ d_{\min}, \, d_{\max} \right] \subset \mathbb{R}^+$ inclusive.
The sensor produces synchronised pairs of colour and depth images $\mbf{C}$ and $\mbf{D}$ respectively.

Each colour image $\mbf{C}$ has a number of corresponding object instance segmentation masks $\mbf{S}_k, \ k \in \left\{1 \dots N_d \right\}$ where $N_d \in \mathbb{N}$ is the number of objects detected in $\mbf{C}$.
\section{Proposed Approach}

\begin{figure}[htb]
	\centering
	\includegraphics[width=0.485\textwidth]{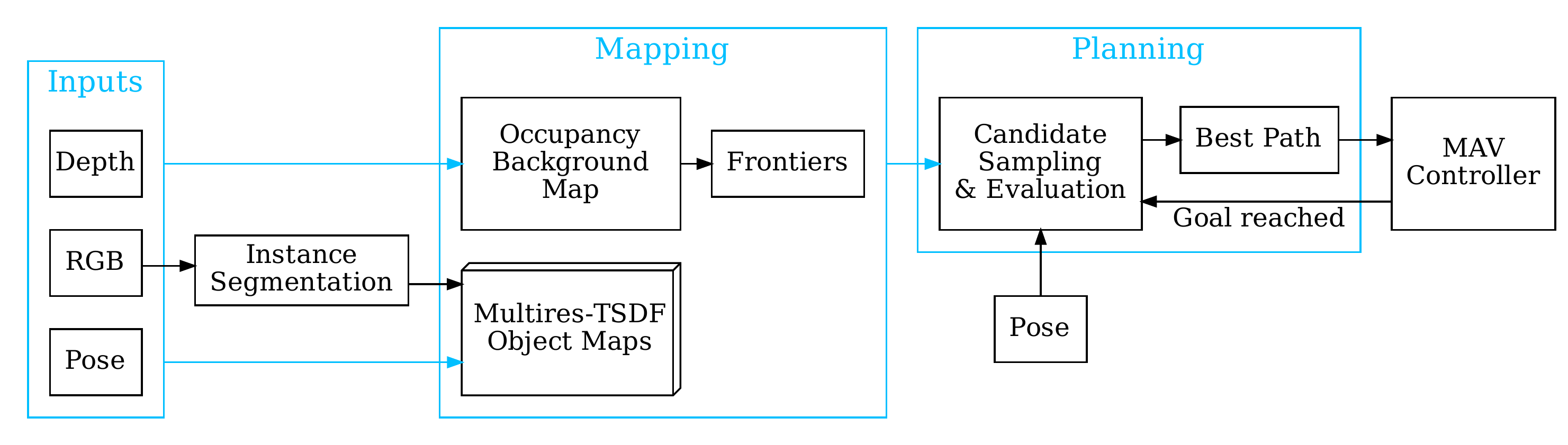}
	\caption{Diagram of the proposed approach. The mapping module receives depth and colour image pairs and the corresponding instance segmentation and pose to update the background and individual object maps and, the set of frontiers. The planning module receives the background and object maps, the set of frontiers and, the current pose to produce the next goal path. Planning happens again when the MAV has completed the goal path.}
	\label{fig_block_diagram}
\end{figure}

Our approach consists of a mapping module and a planning module.
The mapping module receives the current camera pose $\T{W}{C}$, depth and colour images and object instance masks and produces a background map and associated frontiers as well as individual object maps.
The planning module receives the current pose, the background map and its frontiers and the individual object maps and produces the path to the MAV's next goal pose.
The mapping module is run as often as possible using the latest measurements while the planning module is run each time the MAV reaches its goal pose.
Thus, the background and object maps are being continuously updated as the MAV moves to its next goal pose.
The required initial information consists of the semantic classes of interest, their associated resolutions and enough \textit{free} space around the MAV's initial position to allow planning paths.
Figure \ref{fig_block_diagram} shows a diagram of the pipeline.

\subsection{Background Mapping}

The background map with voxel resolution $r \in \mathbb{R}^+$ is created using a modified version of the multi-resolution occupancy mapping pipeline from \cite{2021_RAL_FTPPAL} with some extra information stored in the map compared to \cite{2020_ICRA_DPFTL}.
In \cite{2021_RAL_FTPPAL} an octree-based volumetric occupancy map is created in real-time from posed depth images.
Path planning can be performed directly on the map due to its explicit free space representation and the fact that occupancy data is propagated to higher levels of the octree, allowing for efficient occupancy queries.

\subsubsection{Frontier Detection}
As in \cite{2020_ICRA_DPFTL}, the mapping pipeline is modified to maintain a set of frontiers $\mathcal{F}$.
After each map update, a set of candidate frontiers is generated by forming the union of any previously known frontiers with the set of updated octree nodes.
These candidate frontiers are then checked for frontiers at the finest allocated level to produce the new set of frontiers $\mathcal{F}$.

\subsubsection{Observed Distance Fusion}
An addition compared to \cite{2020_ICRA_DPFTL} is storing the minimum distance each octree leaf node has been observed from.
This allows finding map regions that haven't been observed from a small enough distance as described in Section \ref{sec:candidate_raycasting}.

\subsubsection{Colour Fusion}
The final addition to the map compared to \cite{2020_ICRA_DPFTL} is integrating colour information in each voxel.
The colour update is performed as a weighted average using the same weight as for the occupancy probability.

\subsubsection{Map Update}
The background map $M$ is updated as quickly as possible using pairs of depth and colour images and their corresponding poses without filtering out any detected objects.
This allows performing path planning without having to consider the individual object maps since the background map already contains this information.

\subsection{Object Mapping}

Object mapping was performed using a similar method to \cite{Xu_2019ICRA}, with changes aimed at improving performance, memory usage and reconstruction quality.
We changed the underlying map representation to a modified version of the adaptive-resolution octree-based TSDF mapping from \cite{Vespa_20193DV} which offers improved performance and reduces aliasing artefacts.
A TSDF representation was chosen for objects because of its reduced computational requirements compared to occupancy representations.

\subsubsection{Object Matching}

Similar to \cite{Xu_2019ICRA}, objects detected in the colour image $\mbf{C}$ are matched to known objects in the map, after instance masks smaller than a threshold are discarded to prevent unrealistic detections in synthetic datasets.
Each known object is first raycasted from the current camera pose $\T{W}{C}$ to create a per-object instance mask, taking occlusions from the background and other objects into account.
These raycasted instance masks represent the parts of known objects visible from the current pose.
For each detected object instance mask $\mbf{S}_k$, its intersection over union (IoU) with each per-object raycasted instance mask is computed.
The detected object is associated with the known object for which the highest mask IoU above a certain threshold was achieved.
If there is no IoU above the threshold then a new map is created for the detected object.
While in \cite{Xu_2019ICRA} the object map resolution was constant, in our method it depends on the object's semantic class, allowing efficient mapping of objects of a wide range of sizes.

\subsubsection{Object Map Update}
As in \cite{Xu_2019ICRA}, we integrate colour and foreground probability information in each object map voxel using a weighted average update.
In the case of visible known objects that were undetected in the current colour image, depth and colour information is integrated into them using their raycasted instance masks while omitting the update of foreground probabilities.
Compared to \cite{Xu_2019ICRA}, we also keep track of the minimum observed distance as in the background map.

\subsubsection{Memory Usage Reduction}
We reduce the memory footprint of each voxel by using 16-bit and 8-bit fixed-point numbers for the TSDF value and foreground probability, respectively. This reduces the required memory per voxel by 39\% with negligible degradation of reconstruction accuracy.

\subsection{History of Incomplete Depth Maps}
\label{sec_history}

Depth images often contain invalid or missing data due to e.g.\ reflective surfaces, occlusions or missing geometry in the case of synthetic datasets.
A depth image with a large amount of missing data can cause a naive exploration algorithm to get stuck expecting a reduction in map entropy that never happens due the amount of missing data.
To tackle this issue, we store the history of invalid depth measurements as a low-resolution 3D grid over $V$ containing binary $360^\circ$ images.
All images are initialised to 1 denoting valid depth.
For each depth image $\mbf{D}$ integrated into the background map, the history image $\mbf{H}$ corresponding to pose $\T{W}{C}$ is fetched.
The 2D coordinates of each invalid measurement in $\mbf{D}$ are back-projected in 3D and then re-projected onto $\mbf{H}$, setting the corresponding pixel to 0.

\subsection{Exploration Planning}

Our algorithm is a hybrid between sampling-based and frontier-based exploration.
Candidate next poses are sampled close to the frontiers and the known objects, ranked based on the proposed utility function, and the highest ranked pose is selected as the next goal.
Compared to \cite{2020_ICRA_DPFTL}, in our approach
\begin{itemize}
	\item candidate next positions are also sampled close to objects instead of only close to frontiers,
	\item the exploration utility is computed using two more raycasts in addition to the entropy raycast in \cite{2020_ICRA_DPFTL}, and
	\item the invalid depth history described in Section \ref{sec_history} is used to prevent the exploration from becoming stuck.
\end{itemize}

\subsubsection{Candidate Next Position Sampling}
Candidate next positions $\hat{\mbf{r}}_j \in V, \ j \in \left\{1 \dots n\right\}, \ n \in \mathbb{N}^+$ are sampled among the set of frontiers and the known objects without replacement until $n$ candidates have been sampled or no more frontiers or objects are left to sample, as shown in Figure \ref{fig_exploration} [Left].
Sampling at frontiers favours exploring unknown space while sampling at objects allows obtaining higher quality object observations.
Sampling can be biased towards frontiers or objects using the frontier sampling probability $P_{fr}$ to strike a balance between object reconstruction quality and exploration.

\subsubsection{Path Planning to Candidate Next Positions}
For each sampled candidate position, a path $P_j(\mbf{r}, \hat{\mbf{r}}_j)$ to it is planned from the MAV's current position, as shown in Figure \ref{fig_exploration} [Left].
The path is planned on the background map using the Informed RRT* \cite{Gammell:etal:IROS2014} implementation from the Open Motion Planning Library \cite{Sucan_2012IEEERAM}.
Since sampled positions are at or near non-\textit{free} space, paths are planned as close to them as possible with no requirement for a complete solution.

\subsubsection{Raycasting at Candidate Next Positions}
\label{sec:candidate_raycasting}
In order to evaluate each candidate next position, three low-resolution, $w \times h$, $360^\circ$ raycasts are performed from each candidate position $\hat{\mbf{r}}_j$.
The raycasting resolution $w \times h$ is independent of the input image resolution $W \times H$ and typically much smaller to reduce the computational requirements \cite{Oleynikova_2018IEEERAL}.
This results in three gain images with values in the interval $[0, 1]$ inclusive. The $360^\circ$ raycasting rays from a single candidate are shown in Figure \ref{fig_exploration} [Middle].

An entropy raycast of the background map $M$ is performed, resulting in an entropy gain image $\mbf{G}_{ent}$.
Each pixel of $\mbf{G}_{ent}$ contains the normalised sum of the Shannon entropy of each voxel $\mbf{v}$ along the corresponding ray.
Entropy is accumulated along each ray until $d_{\max}$ or an \textit{occupied} voxel is reached and the sum is normalised with its maximum possible value $\frac{(d_{\max} - d_{\min})}{r}$.
The entropy gain image guides the exploration towards observing \textit{unknown} space resulting in expansion of the map.

A background gain image $\mbf{G}_\mathrm{bg}$ is created by raycasting the minimum observed distance information stored in the background map.
Each pixel of $\mbf{G}_\mathrm{bg}$ contains the distance gain of the octree leaf node $\mbf{n}$ containing the first \textit{occupied} voxel hit by the corresponding ray.
The distance gain of $\mbf{n}$ is
\begin{equation}
	G(\mbf{n}) = \left\{
		\begin{array}{ll}
			0, \ \text{if} \ d_\mathrm{node}(\mbf{n}) \leq d_\mathrm{bg} \vee d_\mathrm{node}(\mbf{n}) \leq d_\mathrm{exp}(\mbf{n}) \\
			\frac{d_\mathrm{node}(\mbf{n}) - \max \left(d_\mathrm{exp}(\mbf{n}), \ d_\mathrm{bg} \right)}{d_{\max}}, \ \text{otherwise}
		\end{array}
		\right.,
		\label{eq_node_gain}
\end{equation}
where $d_\mathrm{node}(\mbf{n}) \in \mathbb{R}^+$ is the minimum distance $\mbf{n}$ has been observed from, $d_\mathrm{exp}(\mbf{n}) \in \mathbb{R}^+$ is the expected observed distance of $\mbf{n}$ from $\hat{\mbf{r}}_j$ and $d_\mathrm{bg} \in \mathbb{R}^+$ is the desired observed distance for the background map.
The desired observed distance $d_\mathrm{bg}$ is set to a value that ensures observed objects will appear large enough in a colour image taken from $d_\mathrm{bg}$ or closer that they can be detected.
The background gain image helps ensure all of the environment has been observed from close enough that all objects have been detected.

Similarly to the background gain image, an object gain image $\mbf{G}_\mathrm{obj}$ is computed by raycasting the minimum observed distance stored in all object maps while taking occlusions into account.
The per-pixel gain is computed in the same manner as for the background except that the object desired observed distance
$d_\mathrm{obj} \in \mathbb{R}^+$ is used instead of $d_\mathrm{bg}$ in Equation \eqref{eq_node_gain}.
The desired observed distance $d_\mathrm{obj} < d_\mathrm{bg}$ for objects of all semantic classes is set to a value that minimises the distance-based depth sensor noise while taking $d_{\min}$ and the MAV dimensions into account.
The object gain image results in the MAV getting closer, higher-quality observations of existing objects.

\subsubsection{Candidate Next Pose Evaluation}
\begin{figure}[t]
	\vspace{0.7em} 
	\centering
	\includegraphics[width=\columnwidth]{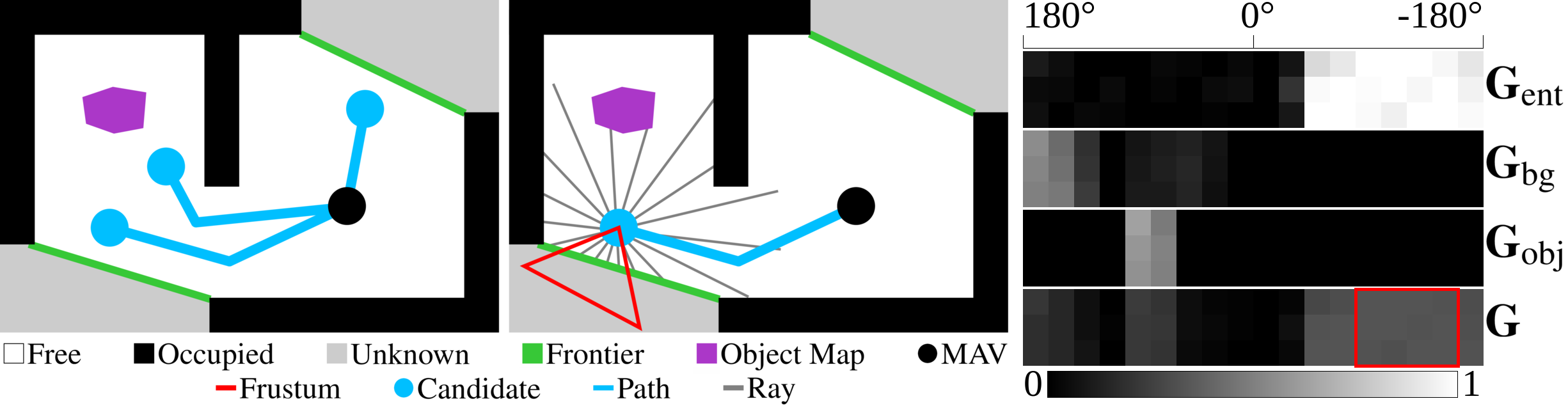}
	\caption{[Left] Candidate view sampling near frontiers and objects, and path planning to candidates from the current pose.
	[Middle] Sparse $360^\circ$ raycasting from one of the candidate views and optimal yaw frustum in red.
	[Right] Entropy, background, object and combined gain images from the raycast shown in [Middle] with the optimal yaw field-of-view in red.}
	\label{fig_exploration}
	\vspace{-0.7em} 
\end{figure}

A single gain image $\mbf{G}$ is produced from a weighted sum of the individual gain images and masking by the invalid depth history
\begin{equation}
	\mbf{G} = \left( \alpha_\mathrm{ent} \mbf{G}_\mathrm{ent} + \alpha_\mathrm{bg} \mbf{G}_\mathrm{bg} + \alpha_\mathrm{obj} \mbf{G}_\mathrm{obj} \right) \circ \ \hat{\mbf{H}},
\end{equation}
where $\alpha_\mathrm{ent}, \alpha_\mathrm{bg}, \alpha_\mathrm{obj} \in \left[0, 1\right], \ \alpha_\mathrm{ent} + \alpha_\mathrm{bg} + \alpha_\mathrm{obj} = 1$ are weighting factors, $\hat{\mbf{H}}$ is the invalid depth history image corresponding to position $\hat{\mbf{r}}_j$ and $\circ$ denotes the element-wise matrix product.
The weighting factors are used to solve the object-level exploration-exploitation dilemma by balancing between exploring, detecting new objects and improving the quality of existing objects.
A sliding window on the gain image $\mbf{G}$ is used to compute the yaw angle $\psi_j$ at which the maximum gain $g_j$ is achieved, similarly to \cite{Selin_2019IEEERAL}.
Some example gain images and the corresponding optimal yaw angle are shown in Figure \ref{fig_exploration} [Right].
The utility $u_j$ of each candidate $j$ is computed to maximise the gain over time as $u_j = g_j\,/\,t_j$,
where the time $t_j$ required for the MAV to complete the path to candidate $j$ is estimated based on the assumption that it always flies at speed $\upsilon_{\max}$ and rotates at speed $\omega_{\max}$.

\subsubsection{Next Goal View}
The candidate $G \in \{1 \dots n\}$ with the highest utility is selected as the next goal.
A yaw angle needs to be assigned to each vertex of the goal path $P_G$. The first vertex, which coincides with the current MAV position, gets assigned the current yaw while the final vertex gets assigned the optimal yaw $\psi_G$. The yaw of intermediate path vertices is computed similarly to the final vertex yaw, by performing $360^\circ$ raycasts and computing the optimal yaw angle.

\subsubsection{Termination Condition}
The exploration stops when $\mathcal{F} = \varnothing$, and all of the background and objects have been observed from a distance at least $d_\mathrm{bg}$ and $d_\mathrm{obj}$, respectively.

\section{Experimental Evaluation and Discussion}

\subsection{MAV Simulator}

To quantitatively evaluate the proposed method and make our results reproducible, we created an MAV simulator suitable for object-centric exploration.
The MAV dynamics are simulated using the prediction step of the MAV Model Predictive Controller from \cite{2020_ICRA_Tzoumanikas}.
The simulator uses Habitat \cite{2019_ICCV_Savva} to provide colour and depth images as well as ground-truth object segmentation masks rendered from the MAV poses.
We included additive Gaussian, zero-mean noise with a standard deviation based on distance in the synthetic depth images to make them more closely resemble those produced by a RealSense D455 sensor.
The noise standard deviation for a depth measurement $d$ was computed as
$\sigma(d) = \max\left(\sigma_{\min}, \ \min\left(s d, \ \sigma_{\max}\right)\right)$.

\subsection{Simulated Experiments}
 
All simulated experiments were run on a computer with an Intel Core i7-6700K CPU, 16 GB of memory and an NVIDIA RTX 3080 GPU.
They were run on Ubuntu 20.04 using ROS Noetic and compiled with GCC 9.4.0 using the O3 optimisation level.
The parameters used for the simulated experiments can be found in Table \ref{table:params}.

\begin{table}[htb]
	\vspace{-0.5em} 
	\centering
	\caption{Simulated experiment parameters}
	\vspace{-0.5em} 
	\label{table:params}
	\begin{tabular}{ | l | c || l | c | }
		\hline
		Parameter & Value & Parameter & Value \\
		\hline
		$\upsilon_{\max}$ & 1.5 m/s
			& $r$, $r_\mathrm{chair}$ & 0.04 m, 0.02 m\\
		$\omega_{\max}$ & 0.75 rad/s & $w \times h$ & $36 \times 10$ \\
		$R$ & 0.125 m
			& $P_{fr}$ & 0.5 \\
		$W \times H$ & $320 \times 240$ 
			& $n$ & 20 \\
		$f_x$, $f_y$ & 262.5 
			& $d_\mathrm{bg}$, $d_\mathrm{obj}$ & 3 m, 1 m \\
		$\sigma_{\min}$, $\sigma_{\max}$ & 0.005 m, 0.2 m
			& $d_{\min}$, $d_{\max}$ & 0.1 m, 10 m \\
		$s$ & 0.002
			& $\alpha_\mathrm{ent}$, $\alpha_\mathrm{bg}$, $\alpha_\mathrm{obj}$ & 0.34, 0.33, 0.33\\
		\hline
	\end{tabular}
	\vspace{-0.5em} 
\end{table}

The proposed semantic exploration method is compared against a classic exploration method, similar to the one from \cite{2020_ICRA_DPFTL} with the addition of object mapping and the invalid depth history described in Section \ref{sec_history}.
The classic exploration method is obtained by setting $\alpha_\mathrm{ent} = 1$, $\alpha_\mathrm{bg} = \alpha_\mathrm{obj} = 0$ and $P_{fr} = 1$.
This is an ablation study of the importance of sampling candidates near objects and the two distance gain images for efficient object-centric exploration.
It also helps showcase that the proposed method retains competitive exploration performance.

\begin{figure}[htb]
	\centering
	\includegraphics[width=\columnwidth]{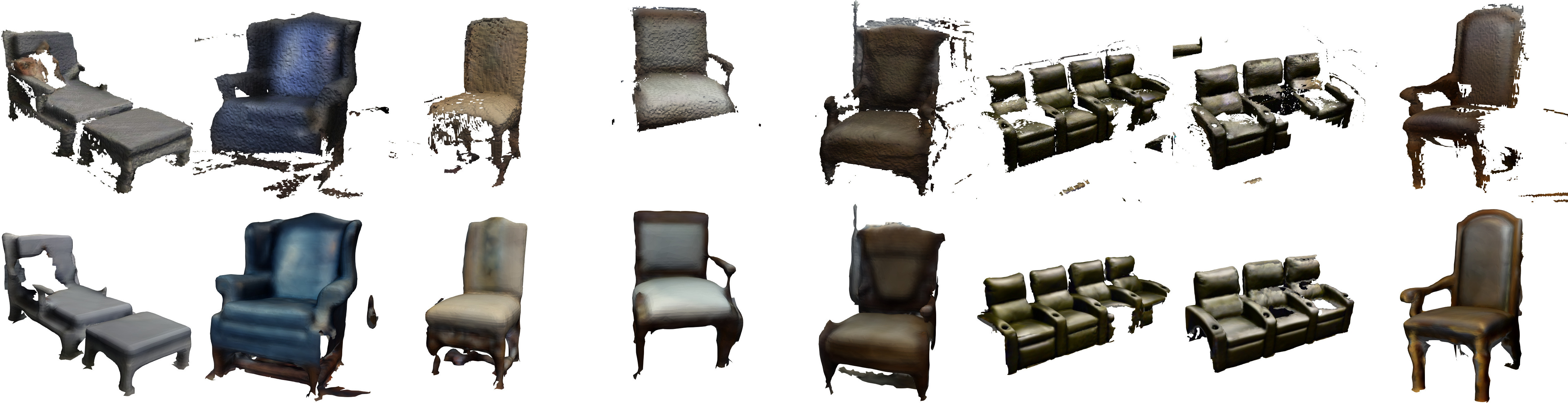}
	\caption{Objects reconstructed using the proposed method [Top] and their corresponding ground-truth meshes [Bottom]. Notice some artefacts due to erroneous segmentation masks.}
	\label{fig_objects}
\end{figure}

Both methods are evaluated on the Matterport3D \cite{2017_3DV_Chang} dataset which consists of semantically annotated 3D scans of real world interior spaces. Since this dataset is collected from real-world data, there is noise and holes in the ground-truth depth and segmentation masks.
The evaluation consists of 5 runs of each method, classic and semantic, on sequences
\texttt{1LXtFkjw3qL},
\texttt{29hnd4uzFmX},
\texttt{2azQ1b91cZZ},
\texttt{2n8kARJN3HM},
\texttt{2t7WUuJeko7},
and
\texttt{8WUmhLawc2A}.
These sequences were chosen because they have relatively little missing data while offering interesting spaces to explore.
Chairs were chosen as the semantic object class of interest due to their abundance in the used sequences.
Some reconstructed objects and their corresponding ground-truth meshes are shown in Figure \ref{fig_objects}.

\begin{figure}[htb]
	\centering
	\includegraphics[width=0.235\textwidth]{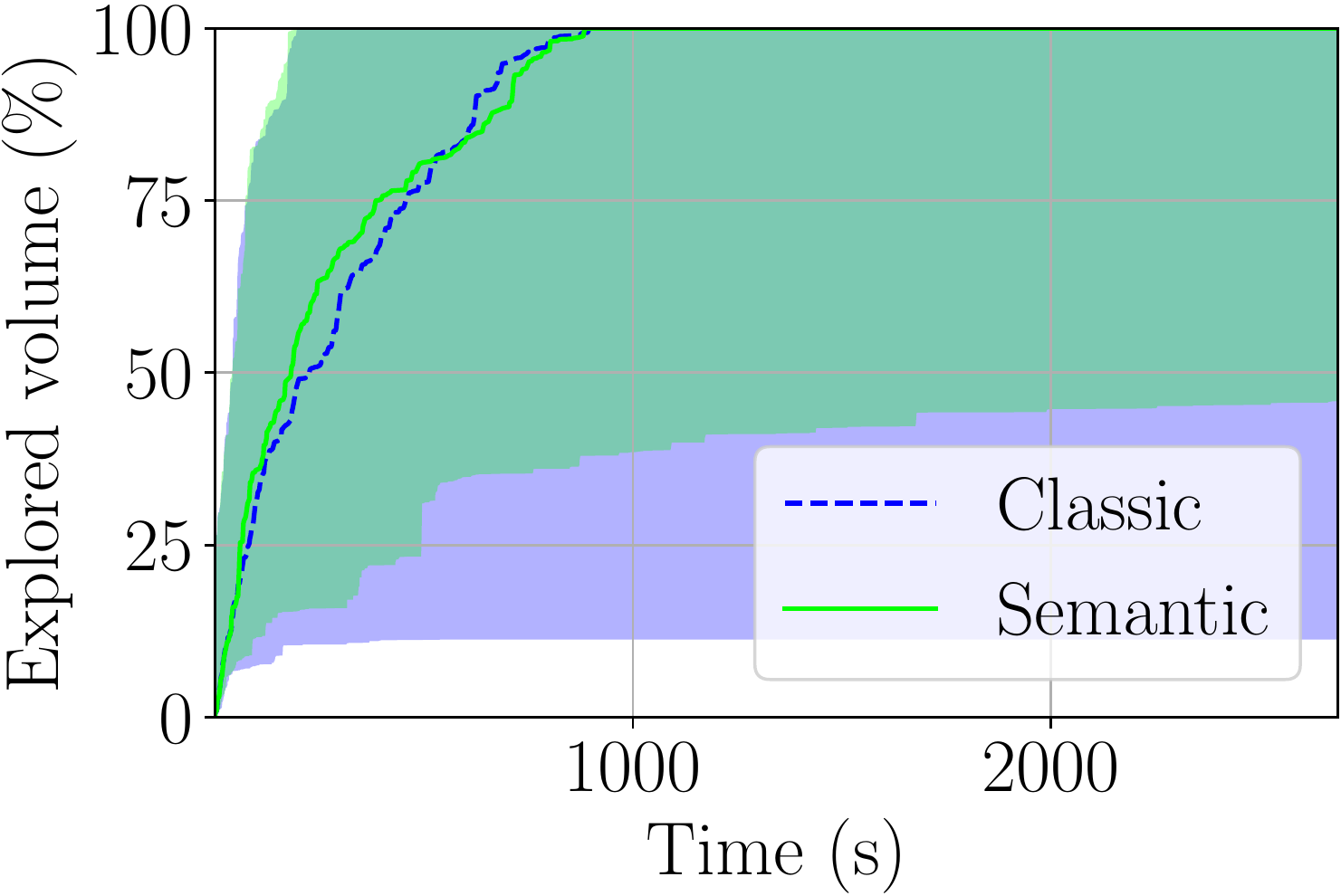}
	\includegraphics[width=0.235\textwidth]{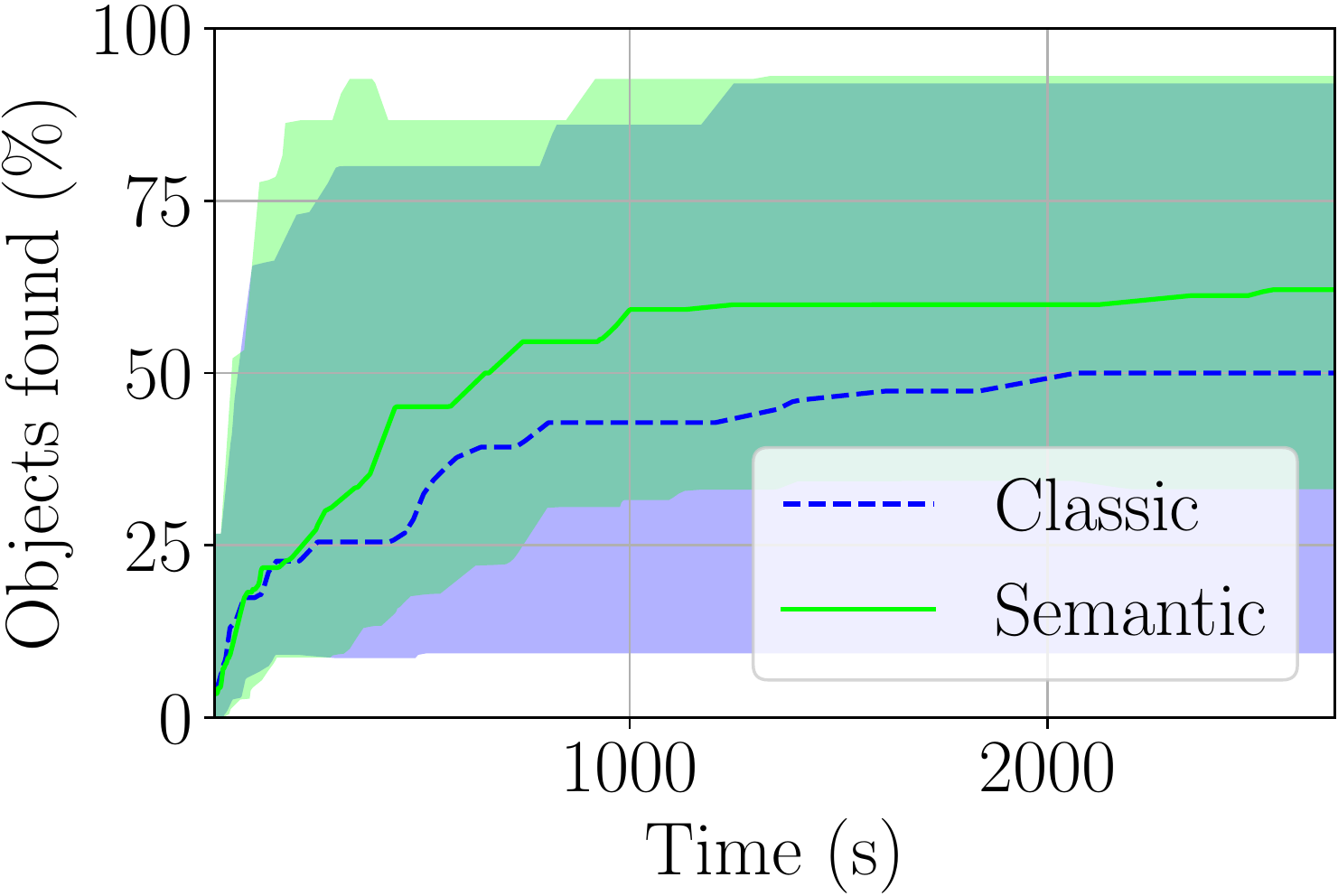}
	\caption{Median, $10^{th}$ and $90^{th}$ percentiles of explored volume [Left] and percentage of objects found [Right].}
	\label{fig_volume_objects}
\end{figure}

Figure \ref{fig_volume_objects} [Left] shows the explored volume over time.
It can be observed that the performance of the proposed method is on-par with classic exploration.

Figure \ref{fig_volume_objects} [Right] shows the percentage of objects detected over time.
The proposed method finds more objects overall, faster and more consistently than the classic approach, indicating the importance of the background distance gain and sampling candidates near objects.
Sampling candidates near known objects can help detect new objects quickly because objects are often close to each other, e.g. chairs tend to be close to other chairs.
One reason for objects not being detected is partial observations, which lead to high uncertainty in the object detection network.
If an object does not cover a large enough part of the image, it cannot be detected.
Thus, observing small parts of an object from different perspectives in different images can result in full exploration while the object remains undetected.
Another problem arises from the imperfect instance segmentation masks in the Matterport3D dataset, where the mask of one object might bleed onto another object.
This can cause the algorithm to erroneously merge the two distinct objects into one, only one of which will be considered matched to a ground-truth object.

\begin{figure}[htb]
	\centering
	\includegraphics[width=0.235\textwidth]{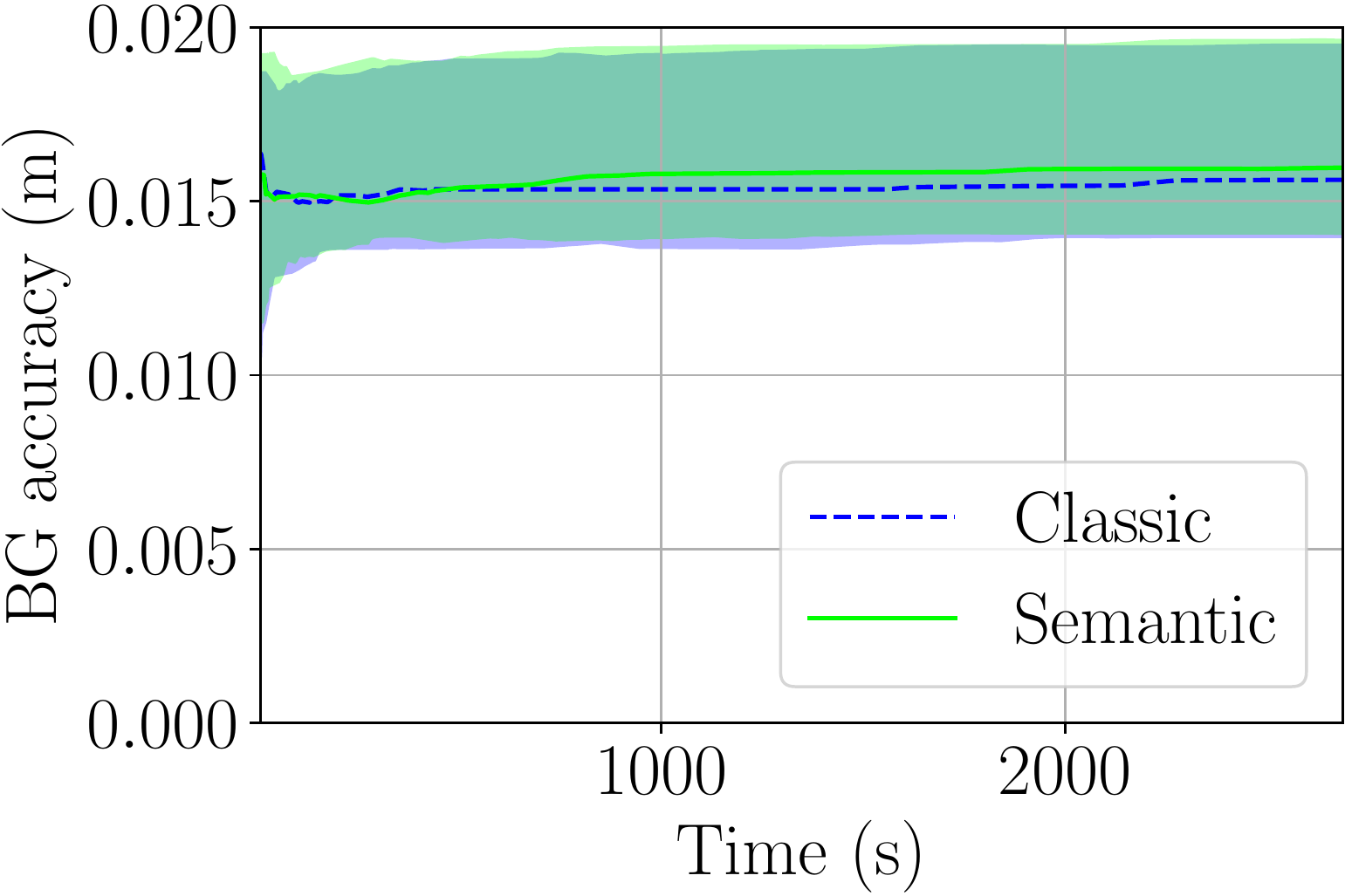}
	\includegraphics[width=0.235\textwidth]{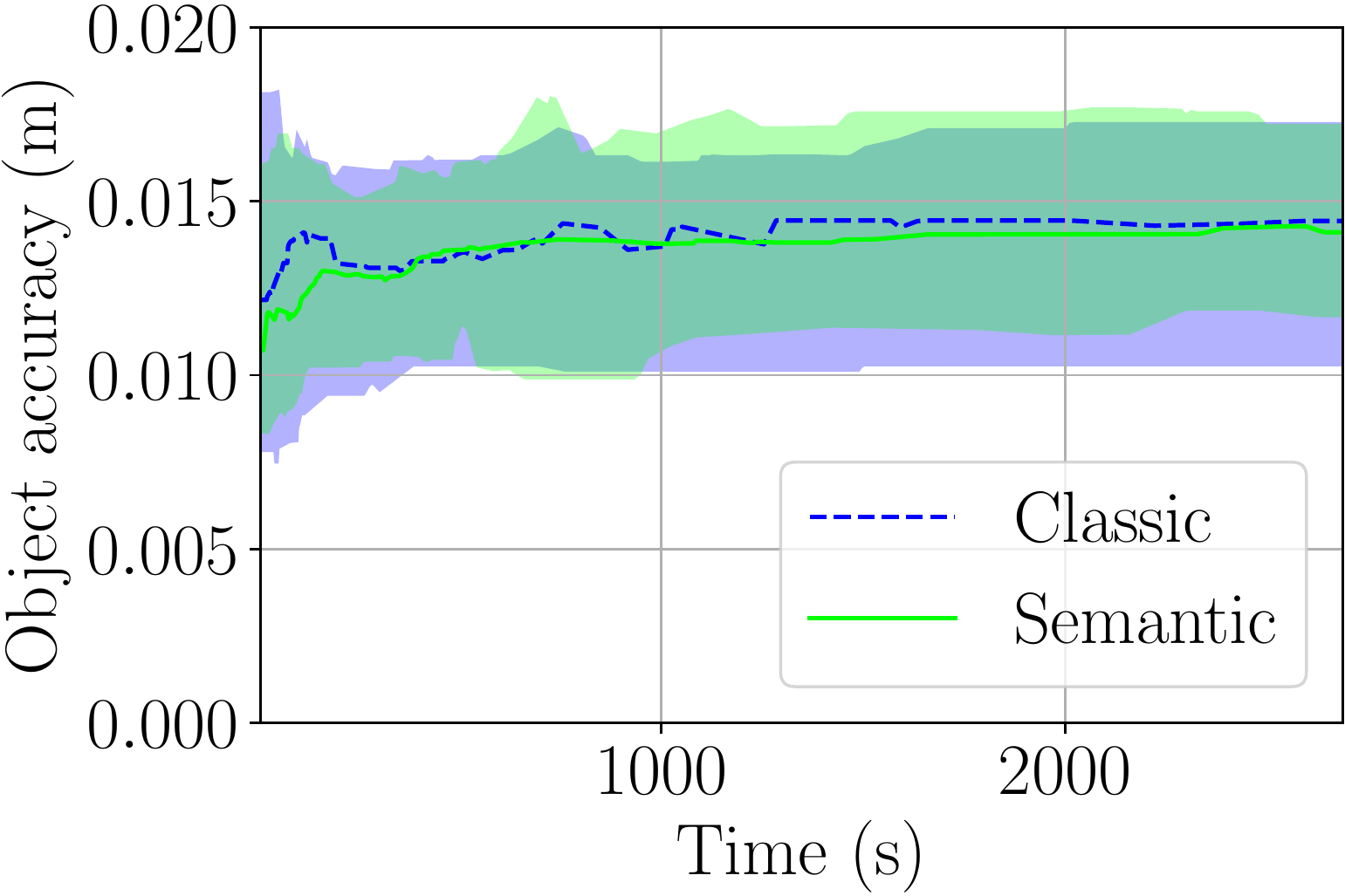}
	\caption{Median, $10^{th}$ and $90^{th}$ percentiles of background [Left] and object [Right] accuracy.}
	\label{fig_accuracy}
\end{figure}

\begin{figure}[htb]
	\centering
	\includegraphics[width=0.235\textwidth]{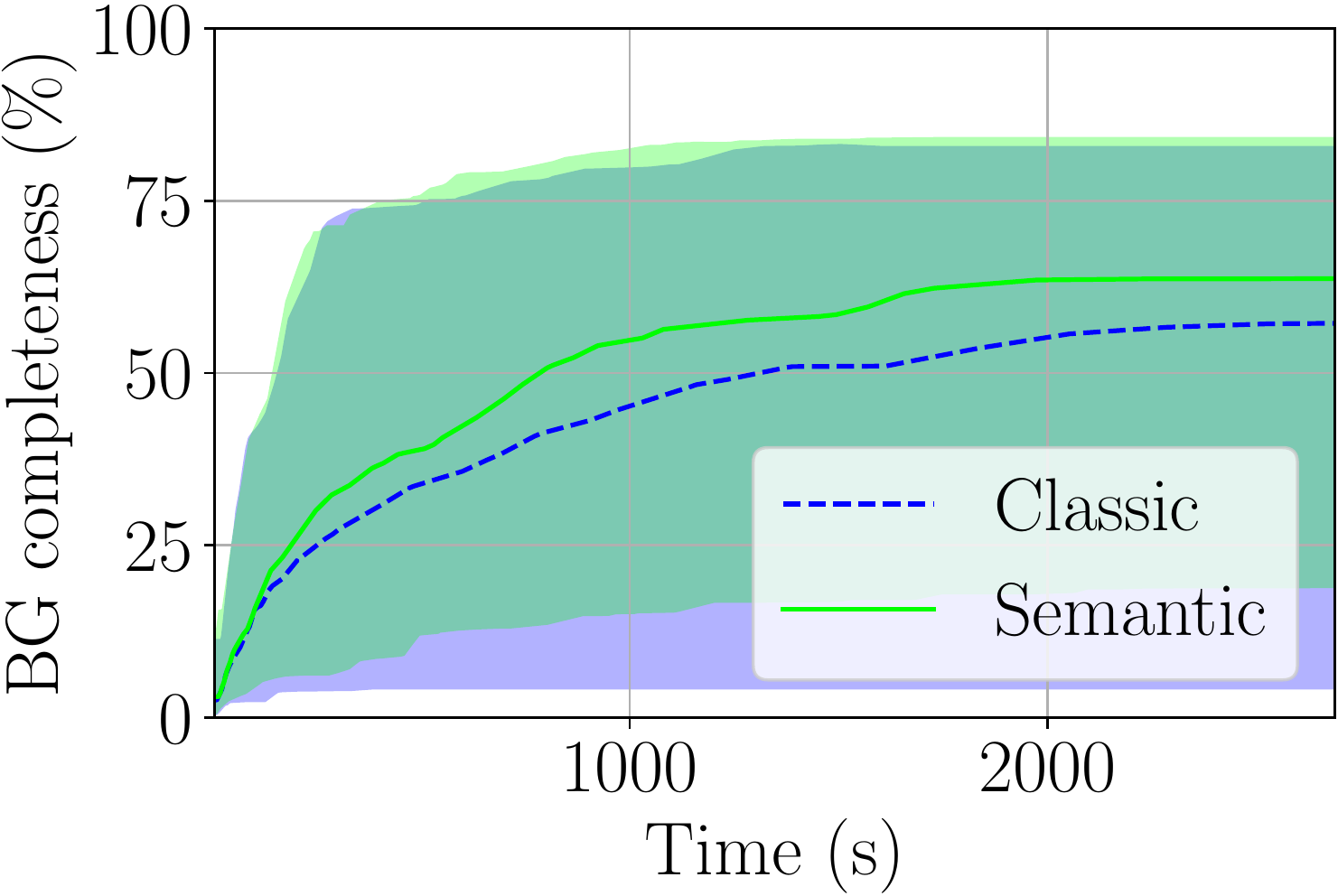}
	\includegraphics[width=0.235\textwidth]{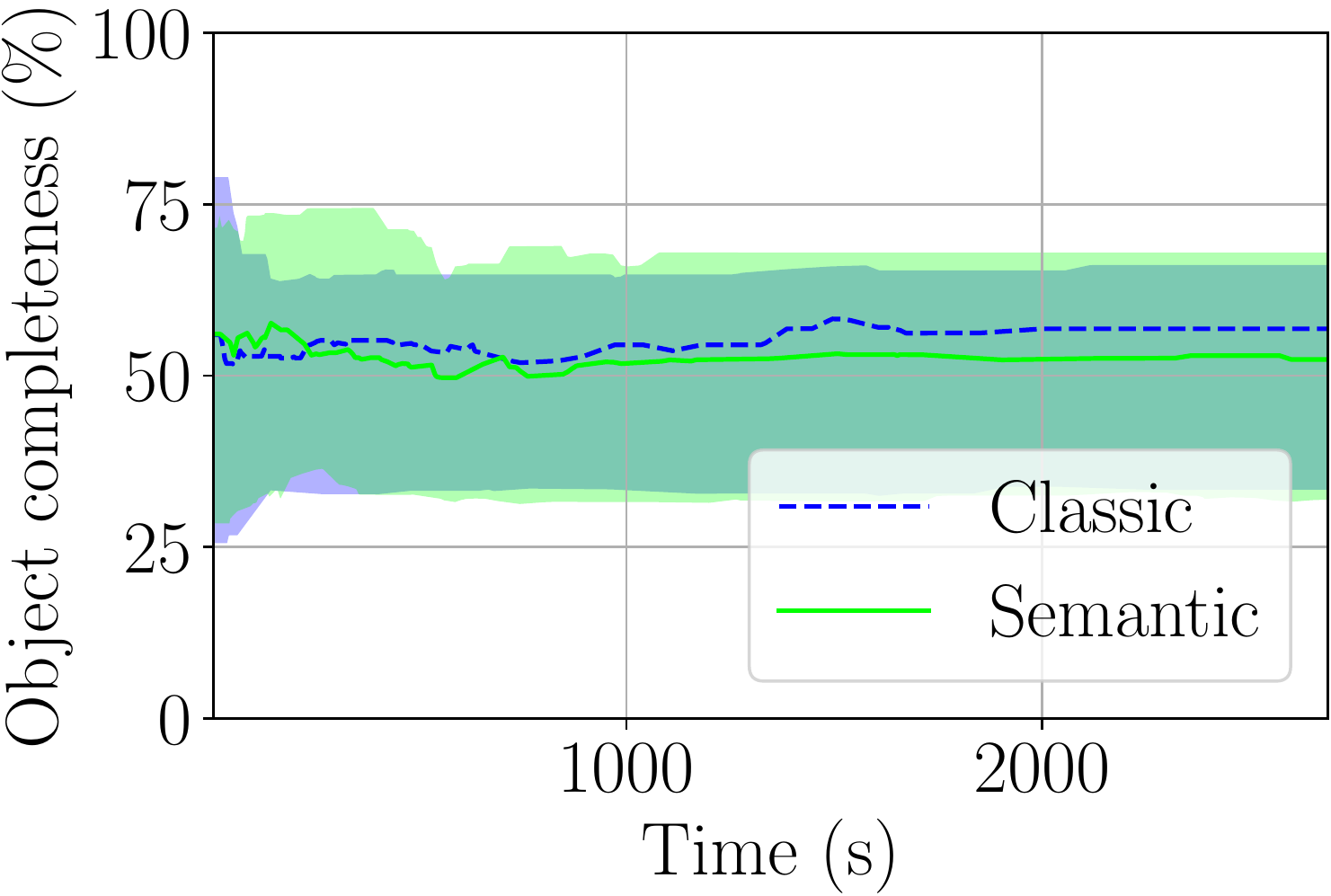}
	\caption{Median, $10^{th}$ and $90^{th}$ percentiles of background [Left] and object [Right] completeness.}
	\label{fig_completeness}
\end{figure}

We use the (pseudo) ground-truth meshes from Matterport3D to evaluate the meshes generated by our method in terms of accuracy and completeness.
Accuracy is computed as the root-mean-square error while completeness is computed as the percentage of the ground-truth mesh vertices for which there is a reconstructed mesh vertex within 5 cm.
Figures \ref{fig_accuracy} and \ref{fig_completeness} show the background and object reconstruction accuracy over time and completeness over time, respectively, aggregated over all runs.
It can be seen that both methods achieve a similar level of reconstruction accuracy.
There is little variation over time in the background and object accuracy because they are observed from a close enough distance even during classic exploration as the Matterport3D consists of interior spaces.
The apparent discrepancy between the explored volume and background completeness is due to the fact that the former is a volume metric while the latter is a surface metric.
This means that the resulting background mesh often has small holes due to occlusions or insufficient data causing a lower completeness score.


\begin{figure}[htb]
	\centering
	\includegraphics[width=0.235\textwidth]{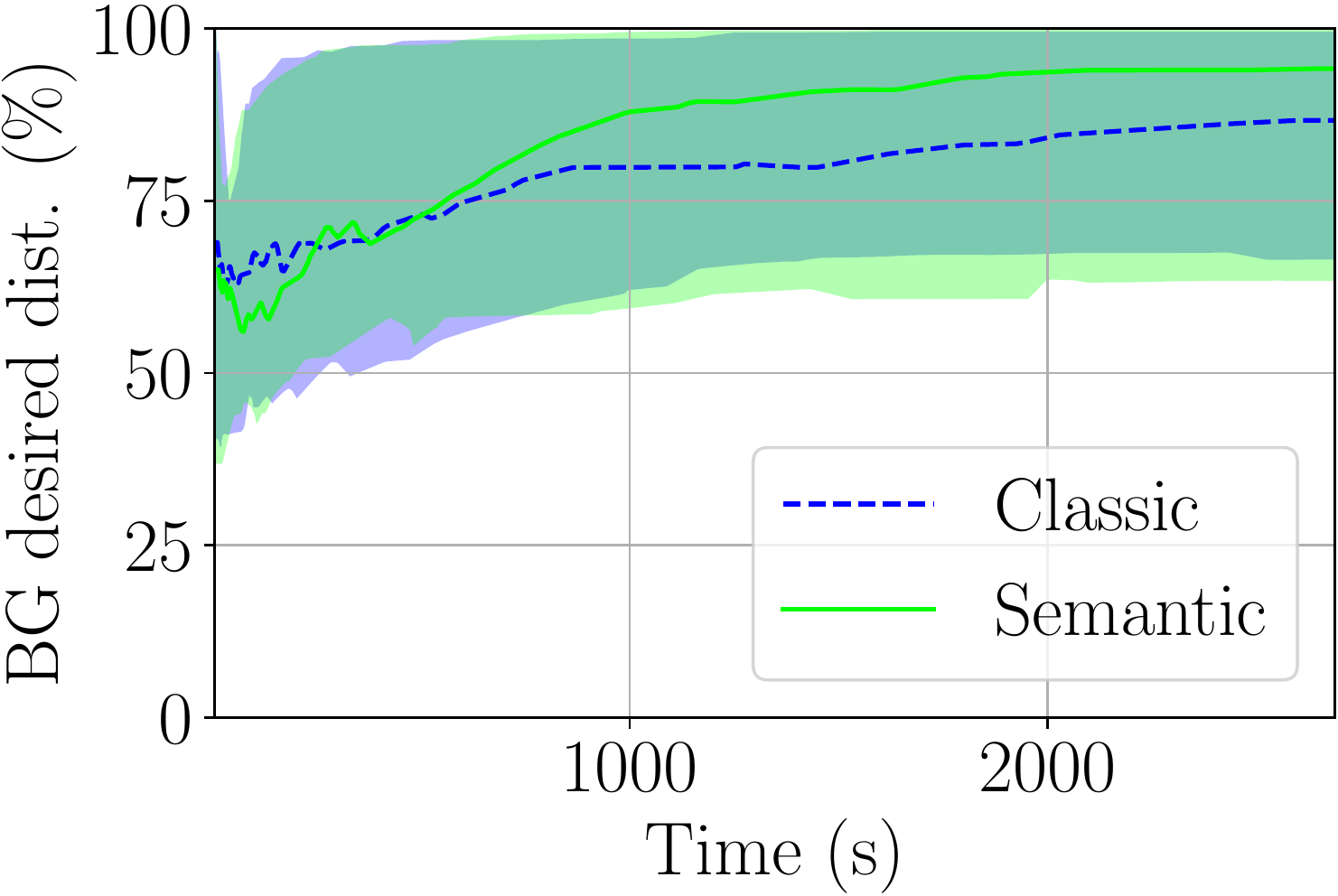}
	\includegraphics[width=0.235\textwidth]{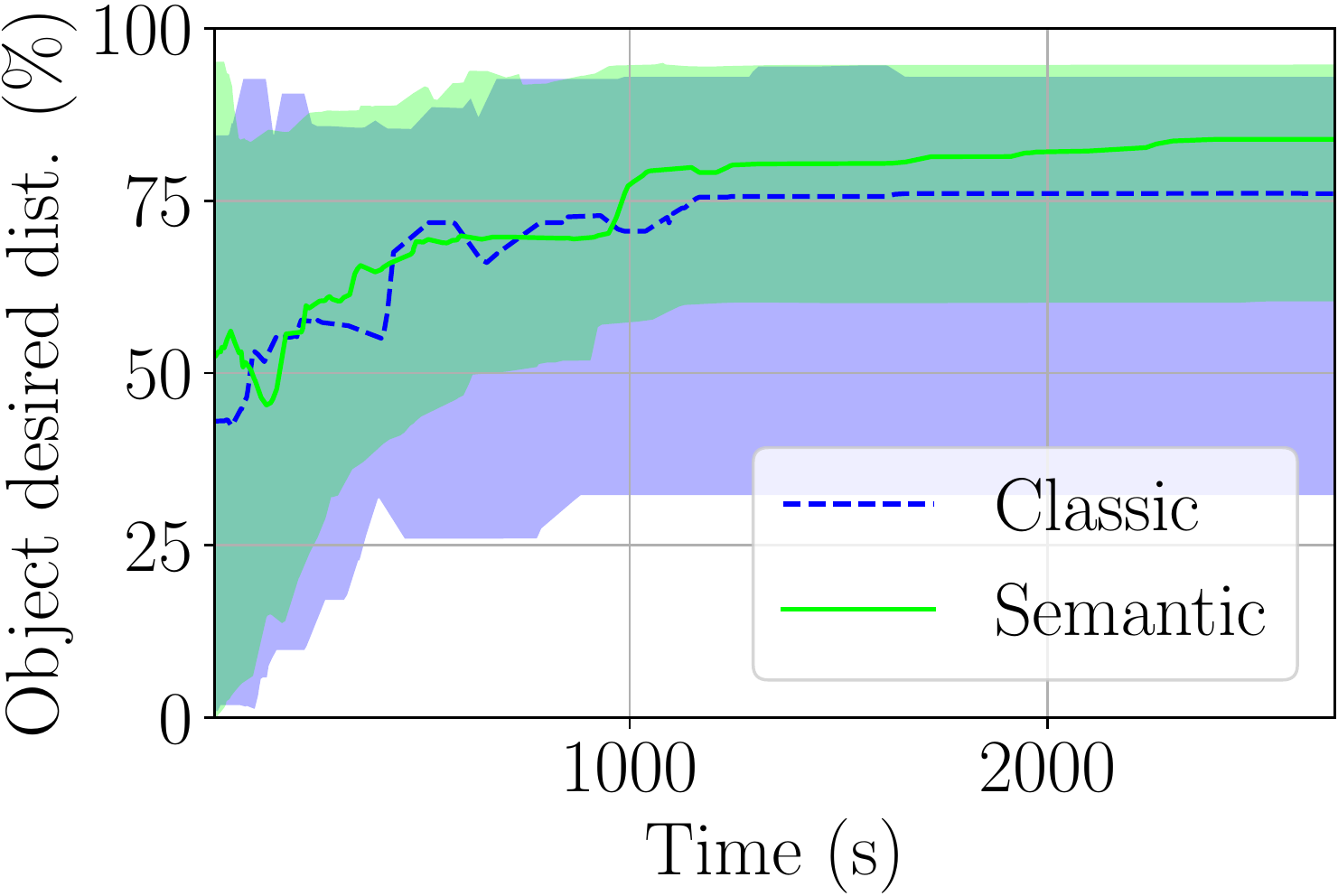}
	\caption{Median, $10^{th}$ and $90^{th}$ percentiles of percentage of background [Left] and objects [Right] observed from at least $d_\mathrm{bg}$ and $d_\mathrm{obj}$ respectively.}
	\label{fig_distance}
\end{figure}

We also evaluate the percentage of the background and objects that were observed from at most the desired distance, $d_\mathrm{bg}$ and $d_\mathrm{obj}$, respectively, as shown in Figure \ref{fig_distance}.
We observe that the proposed method observes a larger part of the environment and objects from the desired distance and in the case of objects, it does so more consistently.
The inaccurate semantic segmentation masks in the Matterport3D dataset can result in artefacts in the object reconstructions, e.g.\ including part of the background.
These artefacts can be occluded, preventing them from being observed and reducing the percentage of objects observed from at most $d_\mathrm{obj}$.

\subsection{Real World Experiments}

A real world experiment was further conducted in order to showcase the feasibility of using the proposed approach on-board an MAV.
The MAV used for the experiments was a DJI F550 hexacopter equipped with an Intel RealSense D455 RGB-D camera.
The MAV pose was provided by a Vicon motion capture system.
All processing was performed on-board the MAV on an NVIDIA Jetson Xavier NX computer.
The experiments were conducted in a 7 m $\times$ 5.5 m $\times$ 5 m room with a 3.6 m $\times$ 3 m $\times$ 1.3 m volume available to the MAV for safety reasons.
The semantic classes used for object reconstruction were limited to backpacks and keyboards.
The background map resolution was 4 cm and the target object resolution was 2 cm for all semantic classes.
The pre-trained Mask R-CNN \cite{He_ICCV2017} model from \cite{2017_GH_MP_MRCNN} was ported to TensorRT to obtain object segmentation masks.
Segmentation was run on the latest colour image not currently being integrated, as fast as possible, on a background thread.

The experiment lasted for 405 s during which 81 frames were integrated and 31 planning iterations took place.
The average frame integration time was 1.835 s, the average segmentation time was 1.167 s and the average planning time was 2.803 s.
Segmentation information was available for 71 frames.
This rather high percentage was achieved because the segmentation time was typically smaller than the integration time so that the segmentation information for the next frame was most often available before the integration of the current frame has finished.
During exploration 2 out of 3 backpacks and 2 out of 3 keyboards placed in the environment were detected even though all of them were observed by the camera.
This is due to the fact that the deployed Mask R-CNN wasn't fine-tuned on this particular environment and didn't detect the objects even though they were visible.

\begin{figure}[htb]
	\centering
	\includegraphics[height=0.105\textheight]{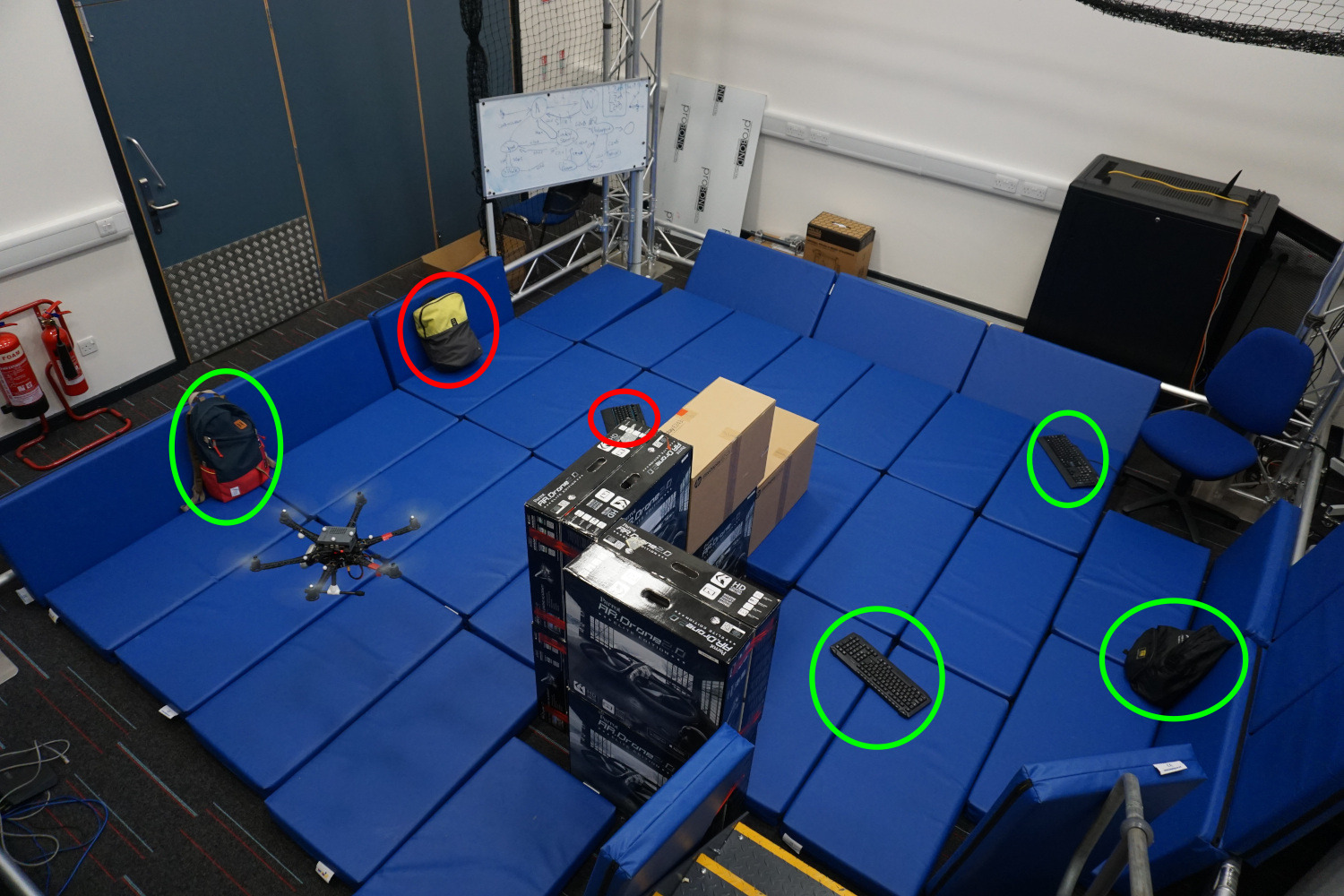}
	\includegraphics[height=0.105\textheight]{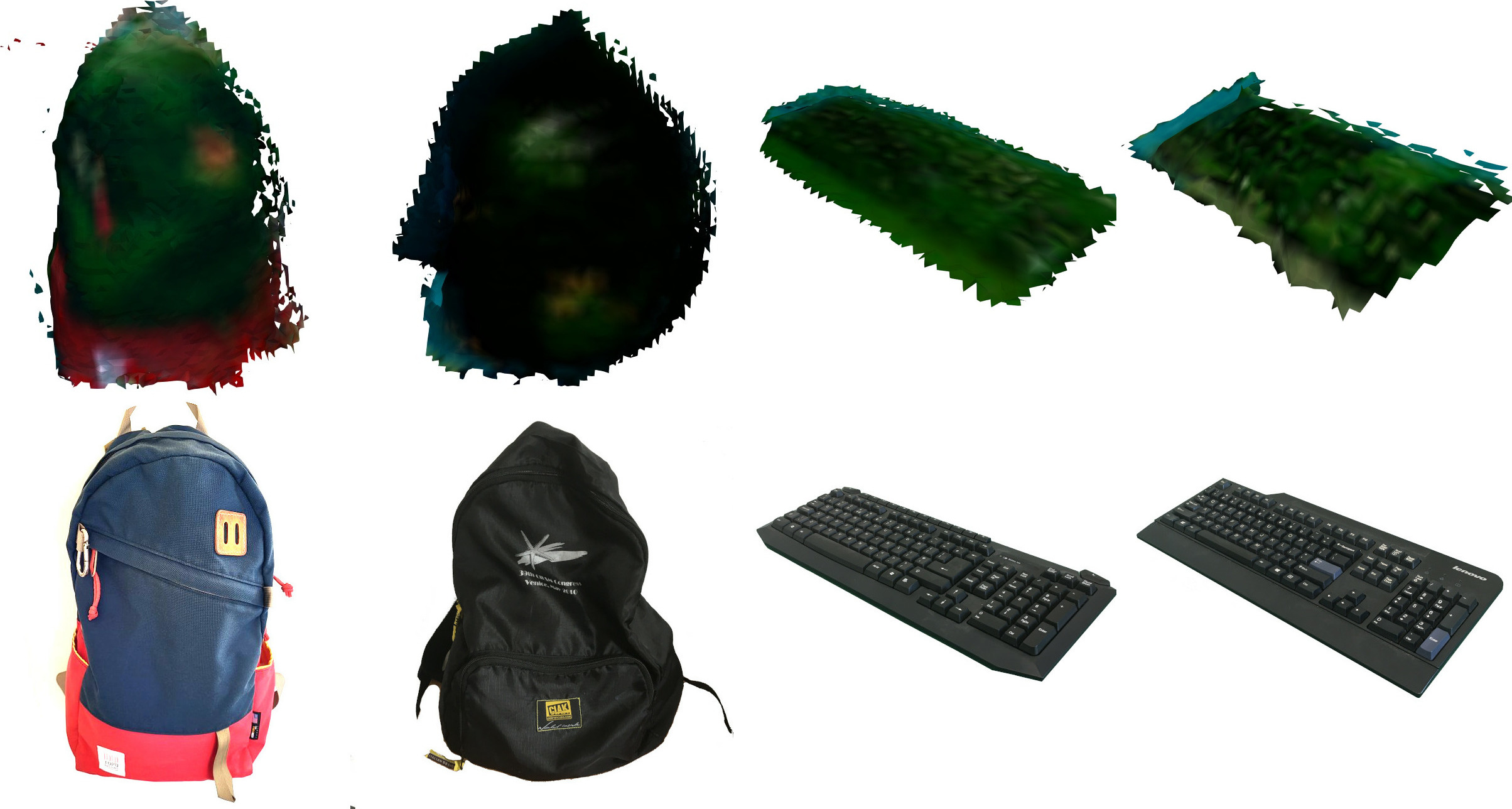}
	\caption{[Left] The real world experimental setup and the MAV.
	Detected and undetected objects are circled in green and red respectively.
	[Top right] Meshes of the 2 backpacks and 2 keyboards reconstructed during the experiment.
	[Bottom right] Photos of the corresponding real world objects.}
	\label{fig:experiment}
	\vspace{-0.3em} 
\end{figure}

\section{Conclusions and Future Work}

In this paper, we have presented a task-specific exploration pipeline for finding and creating high-quality reconstructions of objects in an unknown environment without sacrificing exploration performance.
We have evaluated its effectiveness in extensive simulation studies and have demonstrated that it can be run on-board an MAV in real-world scenes.

We have identified several directions for future work:
more robust object matching, dynamic tracking objects,
object-centric exploration by multiple collaborating robots,
and using machine learning to estimate where undiscovered objects are located given the currently known environment.

\IEEEtriggeratref{17}
\bibliographystyle{IEEEtran}
\input{main.bbl}

\end{document}